\documentclass[11pt]{article}
\usepackage{amsfonts}
\usepackage{mathrsfs}
\usepackage{amsmath,amssymb}
\usepackage{mathtools}
  \usepackage{bm}

\DeclareMathOperator{\EX}{\mathbb{E}}

\usepackage[latin1]{inputenc}
\usepackage{amsmath,amssymb}

\usepackage{amsthm}
\usepackage{latexsym}
\usepackage{epstopdf}
\usepackage{geometry}  
\usepackage{pict2e,picture}
\usepackage{bigints}  
\usepackage[thinlines]{easytable}
\usepackage{enumerate}
\usepackage{array}
\usepackage{fix-cm}
\usepackage{tikz}
\usetikzlibrary{matrix}
\usepackage[linesnumbered,ruled,vlined]{algorithm2e}

\SetCommentSty{mycommfont}

\SetKwInput{KwInput}{Input}                % Set the Input
\SetKwInput{KwOutput}{Output}              % set the Output

\newcolumntype{C}[1]{>{\centering\arraybackslash}m{#1}}
\newcolumntype{R}[1]{>{\raggedleft\arraybackslash}m{#1}}
\usepackage{lipsum}

\usepackage[symbol]{footmisc}

\usepackage{wrapfig}

\usepackage[active]{srcltx}
\usepackage{graphicx}
\usepackage{epstopdf}
\usepackage{enumitem}   
\usepackage[T1]{fontenc}
\usepackage{amsmath}

\usepackage{footnote}
\usepackage{footmisc}
\makesavenoteenv{tabular}
\makesavenoteenv{table}

\usepackage{bbm}

\DeclareMathOperator*{\argmin}{arg\,min}

\usepackage[
bookmarks=true,         %generates bookmarks for all entries in the "Table of Contents"
bookmarksnumbered=true, %includes chapter-/header-/section-/subsection-/... -
colorlinks=true, pdfstartview=FitV, linkcolor=blue, citecolor=blue,
urlcolor=blue]{hyperref}
\usepackage{algpseudocode}

\newtheorem {theorem}{Theorem}[section]
\newtheorem {proposition}{Proposition}[section]

\newtheorem {corollary}{Corollary}[section]

\newtheorem{definition}{Definition}[section]

\newtheorem{lemma}{Lemma}[section]

\newtheorem{remark}{Remark}[section]

\renewcommand\footnotemark{}

\usepackage{dsfont}
\date{\vspace{-5ex}}

\begin{document}

	\title{Nonparametric regression with modified ReLU networks}
	
	\maketitle
	\begin{center}
\bigskip Aleksandr Beknazaryan$^{a}$, Hailin Sang$^{b}$

\bigskip$^{a}$ Institute of Environmental and Agricultural Biology (X-BIO), University of Tyumen\\ Volodarskogo 6, 625003, Tyumen, Russia,
a.beknazaryan@utmn.ru

\bigskip$^{b}$ Department of Mathematics, University of Mississippi, University, MS 38677,  USA
sang@olemiss.edu

 \bigskip
\end{center}
		
	\begin{abstract}
		
		\vskip.2cm 
		
		We consider regression estimation with modified ReLU neural networks in which network weight matrices are first modified by a function $\alpha$ before being multiplied by input vectors. We give an example of continuous, piecewise linear  function $\alpha$ for which the empirical risk minimizers over the classes of modified ReLU networks with $l_1$ and squared $l_2$ penalties attain, up to a logarithmic factor, the minimax rate of prediction of unknown $\beta$-smooth function.
		
		\vskip.2cm \noindent {\bf Keywords}:
		\noindent  convex penalties, $l_p$ regularizations, neural networks, nonparametric regression
		
		\vskip.2cm 
	\end{abstract}
\section{Introduction}

In nonparametric regression estimation we aim to recover an unknown $d$-variate function $g_0$ based on $n$ observed input-output pairs $(\textbf{X}_i, Y_i)\in\mathbb{R}^d\times\mathbb{R}, i=1, ..., n$. Various regression estimating function classes, including wavelets, polynomials, splines and kernel estimates have been studied in the literature (see, e.g.,  \cite{ABS},  \cite{E}, \cite{GM}, \cite{GKKW} and references therein). Along with the development of practical and theoretical applications of neural networks, regression estimations with neural networks are becoming popular in the recent literature (see, e.g., \cite{AB}, \cite{LWLYS}, \cite{LBS}, \cite{LSC}, \cite{OK}, \cite{SH}, \cite{Suz}, \cite{Suz2}, \cite{WCS} and references therein). Usually a class of neural networks with properly chosen architecture and with weight vectors belonging to some regularized set ${\mathcal{W}}_n$ is determined and the estimator $\hat{g}_{ n}$ of $g_0$ is selected to be either the regularized empirical risk minimizer 

\begin{equation}\label{min1}\hat{g}_{ n}\in\argmin\limits_{\textbf{W}\in\mathcal{W}_n}\frac{1}{n}\sum_{i=1}^{n}(Y_i-g_\textbf{W}(\textbf{X}_i))^2\end{equation}
over $\mathcal{W}_n$, or the penalized empirical risk minimizer
\begin{equation}\label{min2}\hat{g}_{ n}\in\argmin\limits_{\textbf{W}}\bigg[\frac{1}{n}\sum_{i=1}^{n}(Y_i-g_\textbf{W}(\textbf{X}_i))^2+J_n(\textbf{W})\bigg],\end{equation}
where $g_\textbf{W}$ is a network with weights $\textbf{W}$ and $J_n$ is some non-negative penalty function. In most cases those two approaches are equivalent in the sense that each of them can be reformulated in the form of the other one by establishing the connection between the regularization of ${\mathcal{W}}_n$ and the penalty $J_n$. Thus, the problem of regression estimation with neural networks can be splitted into 2 major parts:

(i) deriving prediction rates of the empirical risk minimizers \eqref{min1} or \eqref{min2};

(ii) finding an optimization algorithm that identifies the corresponding empirical risk minimizers.

\noindent Convergence rates of empirical risk minimizers (ERM) over the classes of deep ReLU networks are studied in \cite{B1}, \cite{OK}, \cite{SH} and \cite{Suz}. In \cite{B1} it is shown that the ERM of the form \eqref{min1}, with $\mathcal{W}_n$ being the set of weight vectors with coordinates $\{0, \pm1/2, \pm 1, 2\}$, attains, up to logarithmic factors, the minimax rates of prediction of $\beta$-smooth functions. The finiteness of $\mathcal{W}_n$ guarantees that in this case the ERM can be found within finitely many steps.  In \cite{SH} and \cite{Suz} fast rates of prediction are derived for the empirical risk minimizers of the form \eqref{min1} over the sets $\mathcal{W}_n$ of sparse weight vectors. The sparsity assumption is needed to bound the complexity of the class of approximating networks and achieve the desired rates. Penalized empirical risk minimizers of the form \eqref{min2} with the clipped $l_1$ penalty were considered in \cite{OK}. The sparsity induced by the clipped $l_1$ penalty allows to keep the complexity of networks at an appropriate level. However, as the empirical risk, the set of sparse weight vectors of given dimension and the clipped $l_1$ penalty are all non-convex, then, for each of the above sparsity-based approaches, solving the optimization problems \eqref{min1} and \eqref{min2} is very difficult. In fact (see \cite{Nat}), even for the linear systems, the problem of sparse optimization is NP-hard. On the contrary, various techniques have been developed to solve the minimization problems \eqref{min1} and \eqref{min2} for the cases when both the empirical risk and, respectively, the regularization or the penalty are convex (see, e.g., \cite{N}, \cite{PB}, \cite{SST}). However, as for neural networks the empirical risks are always non-convex, the problems \eqref{min1} and \eqref{min2} can not be reduced to a problem of convex optimization,  and the non-convexity in those problems can only be reduced by choosing the set  $\mathcal{W}_n$ or the penalty $J_n$ to be convex. The $l_1$ and $l_2$ network weight regularizations are among the most natural candidates for such choices.
However, in comparison to sparse regularization, those convex constraints may be less convenient for obtaining tight complexity bounds of corresponding classes of regularized networks. For example,  $\delta$-entropy bounds of classes of sparse ReLU networks derived in \cite{SH} depend logarithmically on  $1/\delta$,  while the corresponding bounds for the classes of $l_p$-regularized linear functions and neural networks derived in \cite{B}, \cite{TXL} and \cite{Z} have polynomial dependence on $1/\delta$. Such a strong sensitivity of entropies on the covering radius may not allow to choose the radius to be small enough for achieving minimax rates of convergence. In particular, the order of the error of $l_1$ regularized deep ReLU network estimators obtained in  \cite{TXL} decreases as $1/\sqrt{n}$, which, for big values of $\beta$, is slower than the minimax rate $n^{\frac{-2\beta}{2\beta+d}}$ of prediction of $\beta$-smooth functions. 

In this work we consider a modification of feedforward ReLU networks that is not only determined by the network architecture, the weight matrices and the actions of ReLU activation function, but also by an additional function $\alpha$ that modifies each weight matrix before the latter acts on the input vector. We  give an example of a continuous, piecewise linear  function $\alpha$ for which the modified ReLU networks with the $l_1$ and squared $l_2$ penalties achieve, up to logarithmic factors, the minimax rates of prediction. Thus, the modified neural network models, for which we provide statistical guarantees of convergence to unknown  $\beta$-smooth functions, are the analogs of the well known lasso and ridge methods used in linear regressions models. The tuning parameter for both $l_1$ and squared $l_2$ penalties is equal to $\log_2^6n/n$.

\textit{Notation.} For a function $h:\mathbb{R}\to\mathbb{R}$ and a vector $\textbf{y}=(y_1,...,y_r)\in\mathbb{R}^r$ we denote by $h\cdot\textbf{y}$  the coordinate-wise action of $h$ on $\textbf{y}$: $h\cdot\textbf{y}:=(h(y_1),..., h(y_r)),$ and for a matrix $W\in\mathbb{R}^{p\times r}$ the operation $W\cdot\textbf{y}$ denotes the usual matrix-vector multiplication. The $l_1$ matrix norm (the sum of absolute values of all matrix entries) and the Frobenius norm of the matrix $W\in\mathbb{R}^{p\times r}$ are denoted, respectively, by $\|W\|_1$ and $\|W\|_2$. Also, for a function $g:\Omega\mapsto\mathbb{R}$ and a distribution $P$ on $\Omega$, we denote $\|g\|_{2,P}:=\big(\int_\Omega|g(\textbf{x})|^2dP(\textbf{x})\big)^{1/2}$.

\section{Modified ReLU networks}
For functions $\alpha, \rho:\mathbb{R}\to\mathbb{R}$ and the matrices $V_i\in\mathbb{R}^{p_{i+1}\times p_{i}}$ with $p_0=d$ and $p_{L+1}=1$, let  
$$g(\textbf{x})=V_L\cdot\rho\cdot(\alpha(V_{L-1}))\cdot ...\cdot\rho\cdot (\alpha(V_1))\cdot\rho\cdot(\alpha(V_0))\cdot\textbf{x}, \quad \textbf{x}\in[0,1]^d,$$
be a modified network of depth $L$ and of width $|\textbf{p}|_\infty:=\max\{p_0,...,p_L\}$.
In every hidden layer $i=0, ..., L-1,$ the following three operations are performed:
\begin{itemize}
\item the function $\alpha$ first acts entry-wise on the weight matrix $V_i\in\mathbb{R}^{p_{i+1}\times p_{i}}$ giving a matrix  $\alpha(V_{i})\in\mathbb{R}^{p_{i+1}\times p_{i}}$;
\item the matrix  $\alpha(V_{i})$ is then multiplied by an input vector;
\item the function $\rho$ then acts coordinate-wise on the resulting product.
\end{itemize}
We will consider modified networks determined by the functions 
\begin{equation}\label{alpha}
\alpha(x)=\begin{cases} 
      x+1, & \textrm{if } x<-1, \\
     0, &  \textrm{if } -1\leq x\leq 1, \\
 x-1, & \textrm{if } x>1,
   \end{cases}
\end{equation}
and the ReLU function \begin{align*}
\rho(x)=\begin{cases} 
     0, & \textrm{if } x<0, \\
     x, & \textrm{if } x\geq 0.
   \end{cases}
\end{align*} 
The function $\alpha$ can be considered as a sparsifier that annihilates the weights from the interval $[-1, 1]$. Let
$$\mathcal{G}(L,\textbf{p}):=\{g:[0,1]^d\to\mathbb{R}\; |\;\; g(\textbf{x})=V_L\cdot\rho\cdot(\alpha(V_{L-1}))\cdot ...\cdot\rho\cdot (\alpha(V_1))\cdot\rho\cdot(\alpha(V_0))\cdot\textbf{x}\}$$
be the class of modified ReLU networks of depth $L$ having width vector $\textbf{p}=\{p_0,...,p_{L+1}\}$.

Also, for $F>0$ denote  $\mathcal{G}(L,\textbf{p}, F):=\{g\in\mathcal{G}(L,\textbf{p}); \|g\|_{L^\infty[0,1]^d}\leq F\}$.

Finally, for a given modified network $g(\textbf{x})=V_L\cdot\rho\cdot(\alpha(V_{L-1}))\cdot ...\cdot\rho\cdot (\alpha(V_1))\cdot\rho\cdot(\alpha(V_0))\cdot\textbf{x}$ we denote by 
$$|g|_1:= \sum_{i=0}^L\|V_i\|_1$$
and
$$|g|^2_2:=\sum_{i=0}^L\|V_i\|_2^2$$
 respectively, the $l_1$ norm and the square of the $l_2$ norm of the weights of $g$.
\section{Main results} Consider a nonparametric regression model 
$$Y=g_0(\textbf{X})+\epsilon,$$
where $g_0:[0,1]^d\to\mathbb{R}$ is an unknown regression function to be recovered from $n$ observed iid pairs $(\textbf{X}_i, Y_i)$, $i=1,...,n$. The input variable $\textbf{X}$ follows distribution $P_{\textbf{X}}$ on $[0,1]^d$ and the noise variable $\epsilon$ is assumed to be independent of $\textbf{X}$. It is also assumed that the noise $\epsilon$ is zero mean and sub-Gaussian with $\EX(e^{t\epsilon})\leq  e^{t^2\sigma^2/2}$ for all $t\in\mathbb{R}$ and for some $\sigma>0$, and that $g_0$ belongs to the ball 
\begin{align*}
\mathcal{C}^\beta_d(F):=\bigg\{g:[0,1]^d\to\mathbb{R}: \sum\limits_{0\leq|\boldsymbol{\gamma}|<\beta}\|\partial^{\boldsymbol{\gamma}}g\|_{L^\infty[0,1]^d}+\sum\limits_{|\boldsymbol{\gamma}|=\lfloor\beta\rfloor}\sup\limits_{\substack{\textbf{x},\textbf{y}\in[0,1]^d \\ \textbf{x}\neq \textbf{y}}}\frac{|\partial^{\boldsymbol{\gamma}}g(\textbf{x})-\partial^{\boldsymbol{\gamma}}g(\textbf{y})|}{|\textbf{x}-\textbf{y}|_\infty^{\beta-\lfloor\beta\rfloor}}\leq F\bigg\}
\end{align*} 
of $\beta$-H\"older continuous functions with radius $F$.

Let  $\mathcal{G}(L_n,\textbf{p}_n, F)$ be a class of modified ReLU networks with architecture depending on $n$ and to be specified in Theorem \ref{main} below, and let 
\begin{equation}\label{erm}
\hat{g}_{1, n}\in\argmin\limits_{g\in\mathcal{G}(L_n,\textbf{p}_n, F)}\bigg[\frac{1}{n}\sum_{i=1}^{n}(Y_i-g(\textbf{X}_i))^2+\frac{\log_2^6n}{n}|g|_1\bigg]
\end{equation}
and 
\begin{equation}\label{erm2}
\hat{g}_{2, n}\in\argmin\limits_{g\in\mathcal{G}(L_n,\textbf{p}_n, F)}\bigg[\frac{1}{n}\sum_{i=1}^{n}(Y_i-g(\textbf{X}_i))^2+\frac{\log_2^6n}{n}|g|_2^2\bigg]
\end{equation}
be, respectively, the $l_1$ and the $l_2$ penalized empirical risk minimizers over $\mathcal{G}(L_n,\textbf{p}_n, F)$.
The following theorem estimates the expected prediction error of empirical risk minimizers  \eqref{erm} and \eqref{erm2} with expectation taken over the training data. 
\begin{theorem}\label{main}
	Let $$L_n=8+(\lceil \log_2n\rceil+5)(1+\lceil\log_2(d\lor\beta)\rceil)$$
and $$\normalfont\textbf{p}_n=(d, 6(d+\lceil\beta\rceil)\lceil n^{\frac{d}{2\beta+d}}\rceil, ... , 6(d+\lceil\beta\rceil)\lceil n^{\frac{d}{2\beta+d}}\rceil, 1).$$
	
	Then for the estimators  $\hat{g}_{1, n}$ and  $\hat{g}_{2, n}$ defined in \eqref{erm} and \eqref{erm2} there is a constant $C=C(\beta, \sigma, d, F)$ such that  
	\begin{equation}\label{res}
	\EX\bigg[\|\hat{g}_{1, n}-g_0\|^2_{2,P_{\textbf{X}}}\bigg]\leq Cn^{\frac{-2\beta}{2\beta+d}}\log^7_2n
	\end{equation}
	and
	$$\EX\bigg[\|\hat{g}_{2, n}-g_0\|^2_{2,P_{\textbf{X}}}\bigg]\leq Cn^{\frac{-2\beta}{2\beta+d}}\log^7_2n.$$
\end{theorem}

Thus, the $l_1$ and squared $l_2$ penalized estimators  $\hat{g}_{1, n}$ and  $\hat{g}_{2, n}$ achieve, up to logarithmic factor $\log^7_2n$, the minimax optimal estimation rate $n^{\frac{-2\beta}{2\beta+d}}$ for prediction of $\beta$-smooth functions.

\section{Proofs}

This section is structured as follows:

\begin{itemize}
\item Subsection 4.1 presents the approximating properties and the estimation of entropies of classes of sparse ReLU networks derived in \cite{SH};
\item In Subsection 4.2 we show that while the approximating properties of sparse ReLU networks and $l_1$ regularized ReLU networks are identical, the  entropies of corresponding regularized classes have different dependencies on the imposed constraints;
\item In Subsection 4.3 we establish the connection between the classes of sparse ReLU networks and the classes of modified ReLU networks which allows to derive the approximating capabilities and the estimation of entropies of classes of  $l_1$ and squared $l_2$ regularized modified ReLU networks;
\item Subsection 4.4 provides an oracle inequality for the expected $L_2(P_{\textbf{X}})$ error of penalized estimators;
\item finally, combining the results from Subsections 4.2 and 4.4 we prove Theorem \ref{main}.
\end{itemize}
\subsection{Sparse ReLU networks}

Recall that an ordinary feedforward ReLU network of depth $L$ and of width $|\textbf{p}|_\infty:=\max\{p_0,...,p_L\}$ is a real valued function of the form
$$f(\textbf{x})=W_L\cdot\rho\cdot W_{L-1}\cdot\ ... \cdot\rho\cdot W_1\cdot\rho\cdot W_0\cdot\textbf{x},\quad \textbf{x}\in[0,1]^d,$$ where $W_i\in\mathbb{R}^{p_{i+1}\times p_{i}}$, $i=0,...,L,$ are the weight matrices with $p_0=d$ and $p_{L+1}=1,$ and $\rho(x)=\max\{0, x\}$ is the ReLU activation function. In each hidden layer $i=0,..., L-1,$ the input vector is first multiplied by a weight matrix $W_i$ and the activation $\rho$ then acts  coordinate-wise on the resulting product. Let $$\mathcal{F}(L,\textbf{p}):=\{f:[0,1]^d\to\mathbb{R}\; |\;\; f(\textbf{x})=W_L\cdot\rho\cdot W_{L-1}\cdot\ ... \cdot\rho\cdot W_1\cdot\rho\cdot W_0\cdot\textbf{x}\}$$ denote the class of all ReLU networks of depth $L$ having width vector $\textbf{p}=(p_0, p_1,...,p_{L+1})$. 
For $s>0$ let $\mathcal{F}(L,\textbf{p}, s)$ be the set of sparse networks from $\mathcal{F}(L,\textbf{p})$ having at most $s$ nonzero weights and for $M>0$ let $\mathcal{F}_M(L,\textbf{p}, s)$ be the subset of $\mathcal{F}(L,\textbf{p}, s)$ consisting of sparse networks with all network weights contained in $[-M, M]$.  Choosing $M=1$ the following approximation of functions from $\mathcal{C}^\beta_d(F)$ by sparse networks from $\mathcal{F}_1(L,\textbf{p}, s)$ is given in \cite {SH}:

\begin{theorem}[Schmidt-Hieber, \cite{SH}, Theorem 5]\label{or}
	For any function $f\in\mathcal{C}^\beta_d(F)$ and any integers $m\geq 1$ and $N\geq(\beta+1)^d\lor(F+1)e^d,$ there exists a network $\tilde{f}\in\mathcal{F}_1(L,\emph{\textbf{p}},s)$ with depth
	$$L=8+(m+5)(1+\lceil\log_2(d\lor\beta)\rceil),$$
	width vector $$\emph{\textbf{p}}=(d, 6(d+\lceil\beta\rceil)N, ... ,  6(d+\lceil\beta\rceil)N, 1)$$	
	and number of nonzero parameters
	$$s\leq141(d+\beta+1)^{3+d}N(m+6),$$
	such that 
	$$\|\tilde{f}-f\|_{L^\infty[0,1]^d}\leq(2F+1)(1+d^2+\beta^2)6^dN2^{-m}+F3^\beta N^{-\frac{\beta}{d}}.$$
\end{theorem}
Theorem \ref{or} establishes the approximating capability of the class $\mathcal{F}_1(L,\textbf{p}, s)$. The complexity of the class $\mathcal{F}_1(L,\textbf{p}, s)$ can be determined using the Lipschitz continuity of the ReLU activation function and the sparsity of networks from $\mathcal{F}_1(L,\textbf{p}, s)$. Let us first define the notions of covering numbers and entropies that usually represent complexities of function classes.

\begin{definition}Let $\delta>0$ and let $\mathcal{F}$ be a set of functions from $[0,1]^d$ to $\mathbb{R}$. The $\delta$-covering number $\mathcal{N}(\delta,\mathcal{F},\|\cdot\|_\infty)$ of $\mathcal{F}$ with respect to the $\|\cdot\|_\infty$ distance of functions on $[0,1]^d$ is the minimal number $N\in\mathbb{N}$ such that there exist $f_1,...,f_N$  from $[0,1]^d$ to $\mathbb{R}$ with the property that for any $f\in\mathcal{F}$ there is some $k\in \{1,...,N\}$ such that $$\|f-f_k\|_{L^\infty[0,1]^d}\leq\delta.$$ The number $\log_2\mathcal{N}(\delta,\mathcal{F},\|\cdot\|_\infty)$ is called a $\delta$-entropy of the set  $\mathcal{F}$.
\end{definition}
The following entropy bound of the class $\mathcal{F}_1(L,\textbf{p}, s)$ is obtained in \cite{SH}:

\begin{lemma}[Schmidt-Hieber, \cite{SH}, Lemma 5]\label{ent1}
If $V:=\prod_{i=0}^{L+1}(p_i+1)$, then, for any $\delta>0$,
\begin{equation}\label{e}\log_2\mathcal{N}(\delta, \mathcal{F}_1(L,\normalfont\textbf{p}, s),\|\cdot\|_\infty)\leq (s+1)\log_2\big(2\delta^{-1}(L+1)V^2\big)\leq 4sL\log_2\big(8\delta^{-1}L|\textbf{p}|_\infty\big).\end{equation}
\end{lemma}
Note that the above entropy bound depends
\begin{itemize}
\item linearly on the sparsity $s$, which, according to Theorem \ref{or}, is much smaller than the total number of network weights;
\item linearly, up to a logarithmic factor, on the depth $L$, which, in turn, depends logarithmically on the approximation error given in Theorem \ref{or};
\item logarithmically on the covering radius $\delta$, allowing to take small radii in the oracle inequalities;
\item logarithmically on the networks width  $|\textbf{p}|_\infty$.
\end{itemize}
Due to the above listed properties, combination of Theorem \ref{or} and Lemma \ref{ent1} allows to attain fast convergence rates for sparse neural network regression estimators. We would therefore like to obtain similar bounds for more practicable $l_1$-regularized ReLU networks.

\subsection{$l_1$-regularized ReLU networks}
For $r>0$ let $\mathcal{\widetilde{F}}(L,\textbf{p}, r)$ be the subset of $\mathcal{F}(L,\textbf{p})$ consisting of networks for which the sum of absolute values of all weights (the $l_1$ norm) is bounded by $r$:
$$\mathcal{\widetilde{F}}(L,\textbf{p}, r):=\{f:[0,1]^d\to\mathbb{R}\; |\;\; f(\textbf{x})=W_L\cdot\rho\cdot W_{L-1}\cdot\ ... \cdot\rho\cdot W_1\cdot\rho\cdot W_0\cdot\textbf{x}, \;\;  \sum_{i=0}^L\|W_i\|_1\leq r\}.$$
Note that as the networks from  $\mathcal{F}_1(L,\textbf{p}, s)$ considered in Theorem \ref{or} have at most $s$ nonzero weights all of which are in $[-1, 1]$, then the $l_1$ weight norms of those networks are bounded by $s$. Thus, $\mathcal{F}_1(L,\textbf{p}, s)\subset\mathcal{\widetilde{F}}(L,\textbf{p}, s),$ and, therefore, Theorem \ref{or} also holds for $\mathcal{\widetilde{F}}(L,\textbf{p}, s)$. The next step would be to obtain an entropy bound similar to \eqref{e} for the class of $l_1$-regularized networks $\mathcal{\widetilde{F}}(L,\textbf{p}, s)$. The desired bound would be of the form
\begin{equation}\label{e8}
\log_2\mathcal{N}(\delta, \mathcal{\widetilde{F}}(L,\textbf{p}, s), \|\cdot\|_\infty)\leq C(L,\textbf{p}, s)\log_2\frac{1}{\delta},\end{equation}
with the factor $C(L,\textbf{p}, s),$ that depends  on $L, \textbf{p}$ and $s,$ being much smaller than the total number of weights $\sum_{i=0}^{L}p_ip_{i+1}$ (note that we can always assume that there is no sparsity at all and apply \eqref{e} with $s=\sum_{i=0}^{L}p_ip_{i+1}$, which, however, will result in a highly suboptimal entropy bound).
Suppose that an entropy bound of the form \eqref{e8} holds. For a network $f(\textbf{x})=W_L\cdot\rho\cdot W_{L-1}\cdot\rho\cdot ... \cdot\rho\cdot W_0\cdot\textbf{x}\in\mathcal{\widetilde{F}}(L,\textbf{p}, s)$ denote $\widetilde{W}_i:=s^{-1}W_i$ (divide all entries of $W_i$ by $s$) and define a network $g(\textbf{x})=\widetilde{W}_L\cdot\rho\cdot \widetilde{W}_{L-1}\cdot\rho\cdot ... \cdot\rho\cdot \widetilde{W}_0\cdot\textbf{x}\in\mathcal{\widetilde{F}}(L,\textbf{p}, 1).$ As $\rho$ is positive homogeneous, then
\begin{align*}
&f(\textbf{x})=s^{L+1}g(\textbf{x}).
\end{align*}
Thus,
\begin{align*}
\mathcal{\widetilde{F}}(L,\textbf{p}, s)\subset s^{L+1}\mathcal{\widetilde{F}}(L,\textbf{p}, 1),
\end{align*}
and, therefore, we would have that
\begin{align*}
&\log_2\mathcal{N}(\delta,\mathcal{\widetilde{F}}(L,\textbf{p}, s), \|\cdot\|_\infty)\leq\log_2\mathcal{N}(\delta\big/s^{L+1}, \mathcal{\widetilde{F}}(L,\textbf{p}, 1), \|\cdot\|_\infty)\\
&\leq C(L,\textbf{p}, 1)\log_2\frac{s^{L+1}}{\delta}\leq  LC(L,\textbf{p}, 1)\log_2\frac{s}{\delta}.
\end{align*}
Hence, in this case the entropy bound would only have logarithmic dependence on the $l_1$ regularization. This suggest that no direct analogue of the bound \eqref{e} with linear dependence on the regularization parameter and with logarithmic dependence on the entropy radius exist for the $l_1$-regularized networks. In fact, the classes of networks with $l_1$ norms of all parameters bounded by $1$ were considered in \cite{TXL} and the estimates of their $\delta$-entropies are of polynomial order $1/\delta^2$. Note that the positive homogeneity of ReLU function and the arguments above suggest that weight regularizers allowing to obtain entropy bounds of logarithmic dependence on the covering radius should be invariant under the operations of multiplication and division of weight matrices by positive constants (e.g., as in the case of sparsity constraints). In the following subsection we show that for the proposed modified ReLU networks the $l_1$ weight regularizations are just as well suitable as the sparse regularizations are suitable for the ordinary ReLU networks.

\subsection{Modified ReLU networks}
For $r>0$ let  $\mathcal{G}_1(L,\textbf{p}, r)$ and $\mathcal{G}_2(L,\textbf{p}, r)$ be the subsets of  $\mathcal{G}(L,\textbf{p})$ with, respectively, the $l_1$ norm and the square of the $l_2$ norm of all network weights bounded by $r:$
\begin{align*}
&\mathcal{G}_1(L,\textbf{p}, r):=\\
&\{g:[0,1]^d\to\mathbb{R}\; |\;\; g(\textbf{x})=V_L\cdot\rho\cdot(\alpha(V_{L-1}))\cdot ...\cdot\rho\cdot (\alpha(V_1))\cdot\rho\cdot(\alpha(V_0))\cdot\textbf{x}, \sum_{i=0}^L\|V_i\|_1\leq r\},
\end{align*}
and 
\begin{align*}
&\mathcal{G}_2(L,\textbf{p}, r):=\\
&\{g:[0,1]^d\to\mathbb{R}\; |\;\; g(\textbf{x})=V_L\cdot\rho\cdot(\alpha(V_{L-1}))\cdot ...\cdot\rho\cdot (\alpha(V_1))\cdot\rho\cdot(\alpha(V_0))\cdot\textbf{x}, \sum_{i=0}^L\|V_i\|^2_2\leq r\}.
\end{align*}
Recall also that for $M, s>0$ the class $\mathcal{F}_M(L,\textbf{p}, s)$ consists of ReLU networks with at most $s$ nonzero weights all of which are in $[-M, M]$. The following inclusions hold:
\begin{proposition}\label{inc}
We have
\begin{equation}\label{inc1}\normalfont\mathcal{F}_M(L,\textbf{p}, s)\subset\mathcal{G}_1(L,\textbf{p}, s(M+1))\subset\mathcal{F}_{s(M+1)}(L,\textbf{p}, s(M+1)+|\textbf{p}|_\infty)\end{equation}
 and
\begin{equation}\label{inc2}\normalfont\mathcal{F}_M(L,\textbf{p}, s)\subset\mathcal{G}_2(L,\textbf{p}, s(M+1)^2)\subset\mathcal{F}_{s(M+1)}(L,\textbf{p}, s(M+1)^2+|\textbf{p}|_\infty).\end{equation}
\end{proposition}
\begin{proof}
Take any $f\in\mathcal{F}_M(L,\textbf{p}, s)$ given by $f(\textbf{x})=W_L\cdot\rho\cdot W_{L-1}\cdot\ ... \cdot\rho\cdot W_1\cdot\rho\cdot W_0\cdot\textbf{x}$. Define a function 
\begin{equation}\label{beta}
\nu(x)=\begin{cases} 
      x-1, & \textrm{if } x<0, \\
     0, &  \textrm{if }x=0, \\
 x+1, & \textrm{if } x>0,
   \end{cases}
\end{equation}
and take $V_L=W_L$ and $V_i=\nu(W_i)$, where $\nu$ acts entry-wise on the matrices $W_i, i=0,...,L-1$. As $\alpha(\nu(x))=x,$ $x\in\mathbb{R}$, then $\alpha(V_i)=W_i,$ $i=0,...,L-1$. Therefore, for $g(\textbf{x})=V_L\cdot\rho\cdot(\alpha(V_{L-1}))\cdot ...\cdot\rho\cdot (\alpha(V_1))\cdot\rho\cdot(\alpha(V_0))\cdot\textbf{x}$ we have that $f(\textbf{x})\equiv g(\textbf{x})$. It remains to note that as there are no more than $s$ nonzero entries in the matrices $W_i,$ $i=0,...,L$, all of which are in  $[-M, M]$, then, from the definition of the function $\nu$, we have that 

$$\sum_{i=0}^L\|V_i\|_1=\|W_L\|_1+\sum_{i=0}^{L-1}\|\nu(W_i)\|_1\leq s(M+1)$$
and 
$$\sum_{i=0}^L\|V_i\|_2^2=\|W_L\|_2^2+\sum_{i=0}^{L-1}\|\nu(W_i)\|_2^2\leq s(M+1)^2,$$
which proves the first inclusions in \eqref{inc1} and \eqref{inc2}. To show the second inclusion of \eqref{inc1} take any modified network $g(\textbf{x})=V_L\cdot\rho\cdot(\alpha(V_{L-1}))\cdot ...\cdot\rho\cdot (\alpha(V_1))\cdot\rho\cdot(\alpha(V_0))\cdot\textbf{x}$ from  $\mathcal{G}_1(L,\textbf{p}, s(M+1))$ and consider it as an ordinary ReLU network with weight matrices  $\alpha(V_0), ... ,\alpha(V_{L-1}), V_L$. Note that the regularization $\sum_{i=0}^L\|V_i\|_1\leq s(M+1)$ implies that the weights of $V_0, ... , V_{L-1}, V_L$ are all in $[- s(M+1),  s(M+1)]$ and only at most $s(M+1)$ of them are outside of the interval $[-1,1]$. Therefore, for the function $\alpha$ defined by \eqref{alpha} we have that there are at most $ s(M+1)$ nonzero weights in the matrices $\alpha(V_0), ... ,\alpha(V_{L-1})$ all of which are in $[- s(M+1),  s(M+1)]$. As there are $p_L\leq|\textbf{p}|_\infty$ weights in $V_L$, then $\mathcal{G}_1(L,\textbf{p}, s(M+1))\subset\mathcal{F}_{s(M+1)}(L,\textbf{p}, s(M+1)+|\textbf{p}|_\infty)$. The second inclusion of \eqref{inc2} can be shown similarly.
\end{proof}

\begin{corollary}\label{cor1} Theorem \ref{or} still holds if in its statement the class of sparse ReLU networks $\normalfont\mathcal{F}_1(L,\textbf{p}, s)$ is replaced with either of the classes of $l_1$ and $l_2$ regularized modified ReLU networks $\normalfont\mathcal{G}_1(L,\textbf{p}, 2s)$ and $\normalfont\mathcal{G}_2(L,\textbf{p}, 4s)$.\end{corollary}
\begin{proof}Taking $M=1$ in Proposition \ref{inc} we get that $\normalfont\mathcal{F}_1(L,\textbf{p}, s)\subset\mathcal{G}_1(L,\textbf{p}, 2s)$ and $\normalfont\mathcal{F}_1(L,\textbf{p}, s)\subset\mathcal{G}_2(L,\textbf{p}, 4s)$.
\end{proof}
From the second inclusions of \eqref{inc1} and \eqref{inc2} it follows that in order to estimate the entropies of the classes  $\normalfont\mathcal{G}_1(L,\textbf{p}, 2s)$ and $\normalfont\mathcal{G}_2(L,\textbf{p}, 4s)$ we need to estimate the entropy of the class of sparse ReLU networks $\normalfont\mathcal{F}_M(L,\textbf{p}, s)$. Using the entropy bound of $\normalfont\mathcal{F}_1(L,\textbf{p}, s)$ given in Lemma \ref{ent1} and the positive homogeneity of the ReLU function we get the following 
\begin{lemma}\label{ent2}
For any $\delta>0$
$$\log_2\mathcal{N}(\delta, \mathcal{F}_M(L,\normalfont\textbf{p}, s),\|\cdot\|_\infty)\leq  8sL^2\log_2\big(8\delta^{-1}LM|\textbf{p}|_\infty\big).$$
\end{lemma}
\begin{proof}
Consider a network $f(\textbf{x})=W_L\cdot\rho\cdot W_{L-1}\cdot\ ... \cdot\rho\cdot W_1\cdot\rho\cdot W_0\cdot\textbf{x}\in\mathcal{F}_M(L,\textbf{p}, s)$. There are at most $s$ nonzero weights in the network $f$ all of which are in $[-M, M]$. As the ReLU function is positive homogeneous, that is, $\rho(ax)=a\rho(x)$ for any $a>0$, then  if we denote $\widetilde{W}_i:=M^{-1}W_i$ (divide all entries of $W_i$ by $M$) and define a network $\tilde{f}(\textbf{x})=\widetilde{W}_L\cdot\rho\cdot \widetilde{W}_{L-1}\cdot ... \cdot\rho\cdot \widetilde{W}_1\cdot\rho\cdot \widetilde{W}_0\cdot\textbf{x}\in\normalfont\mathcal{F}_1(L,\textbf{p}, s),$ we get that
\begin{align*}
&f(\textbf{x})=M^{L+1}\tilde{f}(\textbf{x}).
\end{align*}
Thus,
\begin{align*}
\mathcal{F}_M(L,\textbf{p}, s)\subset M^{L+1}\mathcal{F}_1(L,\textbf{p}, s),
\end{align*}
and, therefore, 
$$\log_2\mathcal{N}(\delta,\mathcal{F}_M(L,\textbf{p}, s), \|\cdot\|_\infty)\leq\log_2\mathcal{N}(\delta\big/M^{L+1},\mathcal{F}_1(L,\textbf{p}, s), \|\cdot\|_\infty)\leq 8sL^2\log_2\big(8\delta^{-1}LM|\textbf{p}|_\infty\big),$$
where the last inequality follows from Lemma \ref{ent1}.
\end{proof}
\begin{corollary}\label{cor2}
For any $\delta>0$,
\begin{equation}\label{e1}\normalfont\log_2\mathcal{N}(\delta, \mathcal{G}_1(L,\textbf{p}, 2s),\|\cdot\|_\infty)\leq16(s+|\textbf{p}|_\infty)L^2\log_2\big(16\delta^{-1}Ls|\textbf{p}|_\infty\big)\end{equation}
and 
\begin{equation}\label{e2}\normalfont\log_2\mathcal{N}(\delta, \mathcal{G}_2(L,\textbf{p}, 4s),\|\cdot\|_\infty)\leq 32(s+|\textbf{p}|_\infty)L^2\log_2\big(16\delta^{-1}Ls|\textbf{p}|_\infty\big).\end{equation}
\end{corollary}
\begin{proof}
Taking $M=1$ in Proposition \ref{inc} we get 
$$\mathcal{G}_1(L,\textbf{p}, 2s)\subset\mathcal{F}_{2s}(L,\textbf{p}, 2s+|\textbf{p}|_\infty)$$
 and
$$\mathcal{G}_2(L,\textbf{p}, 4s)\subset\mathcal{F}_{2s}(L,\textbf{p}, 4s+|\textbf{p}|_\infty),$$
which, together with Lemma \ref{ent2}, gives
\begin{align*}&\log_2\mathcal{N}(\delta, \mathcal{G}_1(L,\textbf{p}, 2s),\|\cdot\|_\infty)\leq\log_2\mathcal{N}(\delta, \mathcal{F}_{2s}(L,\textbf{p}, 2s+|\textbf{p}|_\infty),\|\cdot\|_\infty)\\
&\leq 8( 2s+|\textbf{p}|_\infty)L^2\log_2\big(16\delta^{-1}Ls|\textbf{p}|_\infty\big)\leq 16(s+|\textbf{p}|_\infty)L^2\log_2\big(16\delta^{-1}Ls|\textbf{p}|_\infty\big)\end{align*}
and
\begin{align*}&\log_2\mathcal{N}(\delta, \mathcal{G}_2(L,\textbf{p}, 4s),\|\cdot\|_\infty)\leq\log_2\mathcal{N}(\delta, \mathcal{F}_{2s}(L,\textbf{p}, 4s+|\textbf{p}|_\infty),\|\cdot\|_\infty)\\
&\leq 8( 4s+|\textbf{p}|_\infty)L^2\log_2\big(16\delta^{-1}Ls|\textbf{p}|_\infty\big)\leq 32(s+|\textbf{p}|_\infty)L^2\log_2\big(16\delta^{-1}Ls|\textbf{p}|_\infty\big).\end{align*}
\end{proof}

\subsection{Oracle inequality}
Corollaries \ref{cor1} and \ref{cor2} estimate the speeds of approximation and the entropies of classes of regularized modified ReLU networks.
In the next step we establish an oracle inequality that provides the convergence rate of empirical risk minimizers over classes of penalized estimators in terms of their speeds of approximation and entropies. The proof of the oracle inequality in Lemma \ref{oracle} below follows the lines of the proof of Theorem 1 from \cite{OK}. As the original theorem in \cite{OK} is stated for the clipped $l_1$-penalty, in the Appendix section we present the outline of the proof of the oracle inequality in Lemma \ref{oracle} for any non-negative penalty.

Assume that a nonparametric regression model 
$$Y=g_0(\textbf{X})+\epsilon$$ is given with $n$ iid observations  $(\textbf{X}_i, Y_i)$, $i=1,...,n$. The noise $\epsilon$ is sub-Gaussian with $\EX(e^{t\epsilon})\leq e^{t^2\sigma^2/2}$ and is independent of $\textbf{X},$ which, in turn, follows distribution $P_{\textbf{X}}$ on $[0,1]^d$. 
Suppose $g_0$ is bounded with $\|g_0\|_{L^\infty[0,1]^d}\leq F$ (note that no smoothness of $g_0$ is assumed) and let $\mathcal{G}_n$ be a class of functions from  $[0,1]^d$ to $[-F, F]$. For some non-negative penalty function $J_n: \mathcal{G}_n\mapsto\mathbb{R}_+$ let
$$
\hat{g}_n\in\argmin\limits_{g\in\mathcal{G}_n}\bigg[\frac{1}{n}\sum_{i=1}^{n}(Y_i-g(\textbf{X}_i))^2+J_n(g)\bigg]
$$
be the penalized empirical risk minimizer over $\mathcal{G}_n$. For $t>0$ and $j\in\mathbb{N}_0$ define
$$\mathcal{G}_{n,j,t}:=\{g\in\mathcal{G}_n: J_n(g)\leq 2^jt\}.$$   
\begin{lemma}\label{oracle}
If 
\begin{equation}\label{t}t_n\geq \frac{2^{38}\sigma^4(F^2+1)^2\log_2^2n}{n}\end{equation} is such that for any $t\geq t_n$ and for any $\delta\geq 2^jt/8$

\begin{equation}\label{i}\int_{\delta/(2^{11}K_n^2)}^{\sqrt{\delta}}\log_2^{1/2}\mathcal{N}\bigg(\frac{u}{4K_n},\mathcal{G}_{n,j,t}, \|\cdot\|_\infty\bigg)du\leq\frac{c_{\sigma,F}\delta\sqrt{n}}{\log_2 n}\end{equation}
holds for $K_n:=(\sqrt{32\sigma^2}\log_2^{1/2}n)\lor F$ and for some constant $c_{\sigma,F}\leq1/(2^{20}\sigma^2(F^2+1))$, then 
\begin{equation}\label{or_res}\EX\bigg[\|\hat{g}_n-g_0\|^2_{2,P_{\textbf{X}}}\bigg]\leq 2\inf\limits_{g\in\mathcal{G}_n}\bigg[\|g-g_0\|^2_{2,P_{\textbf{X}}}+J_n(g)\bigg]+Ct_n,\end{equation}
where $C=C(\sigma, F)$ is a constant not depending on $n$.
\end{lemma}
\begin{proof}
See Appendix.
\end{proof}

\textit{\textbf{Proof of Theorem \ref{main}}.}
Let $\mathcal{G}_n:=\mathcal{G}(L_n,\textbf{p}_n, F),$ where

$$L_n=8+(\lceil \log_2n\rceil+5)(1+\lceil\log_2(d\lor\beta)\rceil)$$
and
 $$\normalfont\textbf{p}_n=(d, 6(d+\lceil\beta\rceil)\lceil n^{\frac{d}{2\beta+d}}\rceil, ... , 6(d+\lceil\beta\rceil)\lceil n^{\frac{d}{2\beta+d}}\rceil, 1)$$
 with
\begin{equation}\label{p_n}|{\textbf{p}}_n|_\infty=6(d+\lceil\beta\rceil)\lceil n^{\frac{d}{2\beta+d}}\rceil.\end{equation}	
Applying Corollary \ref{cor1} with $m=\lceil \log_2n\rceil$ and $N=\lceil n^{\frac{d}{2\beta+d}}\rceil$ we get
\begin{equation}\label{1}\inf\limits_{g\in\mathcal{G}_n}\bigg[\|g-g_0\|^2_{2,P_{\textbf{X}}}+\frac{\log_2^6n}{n}|g|_1\bigg]\leq cn^{\frac{-2\beta}{2\beta+d}}\log^7_2n\end{equation}
and
$$\inf\limits_{g\in\mathcal{G}_n}\bigg[\|g-g_0\|^2_{2,P_{\textbf{X}}}+\frac{\log_2^6n}{n}|g|_2^2\bigg]\leq cn^{\frac{-2\beta}{2\beta+d}}\log^7_2n$$
for some constant $c=c(\beta, d, F)$. Take 
\begin{equation}\label{t_n}t_n:=\frac{8|\textbf{p}_n|_\infty\log_2^6n}{n}.\end{equation}
 Obviously, condition \eqref{t} is satisfied for sufficiently large $n$. We now need to check that the condition \eqref{i} is also satisfied for penalties 
$$J'_{ n}(g)=\frac{\log_2^6n}{n}|g|_{1}$$
and $$J''_{ n}(g)=\frac{\log_2^6n}{n}|g|^2_{2}.$$
As the derivations for those two penalties are identical, we will only check that the condition \eqref{i} is satisfied for the penalty $J'_{ n}(g)$. For this penalty we have that 
$$\mathcal{G}_{n,j,t}=\{g\in\mathcal{G}_n: J'_n(g)\leq 2^jt\}=\mathcal{G}_1(L_n,\textbf{p}_n, 2^jnt\log_2^{-6}n).$$   
Applying Corollary \ref{ent2}, we get that for any $t\geq t_n$ and any $\delta\geq 2^jt/8\geq (|\textbf{p}_n|_\infty\log_2^{6}n)/n$

\begin{align*}
&\int_{\delta/(2^{11}K_n^2)}^{\sqrt{\delta}}\log_2^{1/2}\mathcal{N}\bigg(\frac{u}{4K_n},\mathcal{G}_{n,j,t}, \|\cdot\|_\infty\bigg)du\\
&=\int_{\delta/(2^{11}K_n^2)}^{\sqrt{\delta}}\log_2^{1/2}\mathcal{N}\bigg(\frac{u}{4K_n},\mathcal{G}_1(L_n,\textbf{p}_n, 2^jnt\log_2^{-6}n), \|\cdot\|_\infty\bigg)du\\
&\leq 4L_n(2^jnt\log_2^{-6}n+|\textbf{p}_n|_\infty)^{1/2}\int_{\delta/(2^{11}K_n^2)}^{\sqrt{\delta}}\log_2^{1/2}\frac{16L_n|\textbf{p}_n|_\infty2^jnt\log_2^{-6}n}{(u/4K_n)}du\\
&\leq
4L_n(9\delta n\log_2^{-6}n)^{1/2}\sqrt{\delta}\log_2^{1/2}\frac{2^{17}K_n^3L_n|\textbf{p}_n|_\infty 2^jnt\log_2^{-6}n}{\delta}\\
&\leq 12L_n\delta\sqrt{n}\log_2^{-3}n\log_2^{1/2}\big(2^{20}K_n^3L_n|\textbf{p}_n|_\infty n\log_2^{-6}n\big)\leq c\delta\sqrt{n}\log^{-3/2}_2n,
\end{align*}
where the last inequality holds for $K_n\leq\sqrt{32\sigma^2}\log_2^{1/2}n$ and for some constant $c=c(\beta, \sigma, d, F)$. As for any positive constant $c_{\sigma,F}\leq1/(2^{20}\sigma^2(F^2+1))$ and for $n$ sufficiently large we have that 
$$ c\delta\sqrt{n}\log^{-3/2}_2n\leq \frac{c_{\sigma,F}\delta\sqrt{n}}{\log_2 n},$$ then the condition \eqref{i} of Lemma \ref{oracle} is satisfied. The result now follows from equalities and inequalities \eqref{or_res}-\eqref{t_n}.

\begin{remark} Note that in Theorem \ref{main} we only assume that the unknown function $g_0$ belongs to the ball $\mathcal{C}^\beta_d(F)$ and achieve, up to a logarithmic factor, the minimax optimal estimation rate $n^{\frac{-2\beta}{2\beta+d}}$. Under additional composition assumptions on $g_0,$ faster prediction rates of sparse ReLU networks are derived in \cite{SH}. Combining the above proofs with the resuts obtained in \cite{SH}, under the same additional structural assumptions on $g_0,$ identical faster rates of prediction can also be derived for $l_1$ and squared $l_2$ regularized modified ReLU networks.  
\end{remark}
\section{Appendix}
Proof of Lemma \ref{oracle} is based on the following result given in  \cite{GKKW}, Theorem 19.3:

\begin{theorem}\label{t1}
	Let $Z_1,...,Z_n$ be independent and identically distributed random variables with values in $\mathbb{R}^d$. Let $K_1, K_2\geq 1$ and let $\mathcal{F}$ be a class of functions $f:\mathbb{R}^d\to \mathbb{R}$ with the properties $\|f(z)\|_\infty\leq K_1$ and $\EX\{f(Z)^2\}\leq K_2\EX\{f(Z)\}$. Let $0<\omega<1$ and $t^\star>0$. Assume that 
	\begin{equation}\label{cond1}\sqrt{n}\omega\sqrt{1-\omega}\sqrt{t^\star}\geq 288\max\{2K_1, \sqrt{2K_2}\}\end{equation}
	and for any $\delta\geq t^\star/8$,
	\begin{equation}\label{cond2}
	\frac{\sqrt{n}\omega(1-\omega)\delta}{96\sqrt{2}\max\{K_1, 2K_2\}}\geq
	\int_{\frac{\omega(1-\omega)\delta}{16\max\{K_1, 2K_2\}}}^{\sqrt{\delta}}\log_2^{1/2}\mathcal{N}(u, \mathcal{F}, \|\cdot\|_\infty)du.
	\end{equation}
	Then 
	$$\mathbb{P}\bigg(\sup\limits_{f\in\mathcal{F}}\frac{\EX\{f(Z)\}-\frac{1}{n}\sum_{i=1}^nf(Z_i)}{t^\star+\EX\{f(Z)\}}>\omega\bigg)\leq 60 \exp\bigg(-\frac{nt^\star\omega^2(1-\omega)}{128\cdot 2304\max\{K_1^2,K_2\}}\bigg).$$
\end{theorem}

We will follow the strategy used in \cite{OK}. Throughout the proof, for two sequences $a_n$ and $b_n$ the notation $a_n\lesssim b_n$ indicates the existence of a constant $c=c(\sigma, F),$ which possibly depends on $\sigma$ and $F$ but is independent on $n$, such that $a_n\leq cb_n$ for all $n\in\mathbb{N}$. Also,  $\textrm{P}_n$ denotes the empirical distribution based on the data  $(\textbf{X}_i, Y_i)$, $i=1,...,n,$ and the abbreviations $\textrm{P}f:=\int fdP_{\textbf{X}}$ and $\textrm{P}_nf:=\int fdP_n$ are used throughout the proof.

\textit{Proof of Lemma \ref{oracle}}.
Denote $K_n:=(\sqrt{32\sigma^2}\log_2^{1/2}n)\lor F$ and $Y^\dagger:=\textrm{sign}(Y)(|Y|\land K_n)$ and let 
$$g^\dagger(\textbf{x}):=\EX (Y^\dagger |\textbf{X}=\textbf{x})$$
be the regression function of $Y^\dagger$. The proof is based on the decomposition 
$$\|\hat{g}_n-g_0\|_{2,P_{\textbf{X}}}^2=\textrm{P}(Y-\hat{g}_n(\textbf{X}))^2-\textrm{P}(Y-g_0(\textbf{X}))^2=\sum_{i=1}^{4}A_{i,n},$$
with
$$A_{1,n}:=\bigg[\textrm{P}(Y-\hat{g}_n(\textbf{X}))^2-\textrm{P}(Y-g_0(\textbf{X}))^2\bigg]-\bigg[\textrm{P}(Y^\dagger-\hat{g}_n(\textbf{X}))^2-\textrm{P}(Y^\dagger-g^\dagger(\textbf{X}))^2\bigg],$$
$$A_{2,n}:=\bigg[\textrm{P}(Y^\dagger-\hat{g}_n(\textbf{X}))^2-\textrm{P}(Y^\dagger-g^\dagger(\textbf{X}))^2\bigg]-2\bigg[\textrm{P}_n(Y^\dagger-\hat{g}_n(\textbf{X}))^2-\textrm{P}_n(Y^\dagger-g^\dagger(\textbf{X}))^2\bigg]-2J_n(\hat{g}_n),$$
$$A_{3,n}:=2\bigg[\textrm{P}_n(Y^\dagger-\hat{g}_n(\textbf{X}))^2-\textrm{P}_n(Y^\dagger-g^\dagger(\textbf{X}))^2\bigg]-2\bigg[\textrm{P}_n(Y-\hat{g}_n(\textbf{X}))^2-\textrm{P}_n(Y-g_0(\textbf{X}))^2\bigg]$$
and
$$A_{4,n}:=2\bigg[\textrm{P}_n(Y-\hat{g}_n(\textbf{X}))^2-\textrm{P}_n(Y-g_0(\textbf{X}))^2\bigg]+2J_n(\hat{g}_n).$$

Note that the terms $A_{1,n}$ and $A_{3,n}$ above do not depend on the penalty $J_n$, and, using the sub-Gaussianity of the noise $\epsilon$, it is shown in \cite{OK}, page 18, that
$$A_{1,n}\lesssim\frac{\log_2n}{n} \quad \textrm{and} \quad \EX(A_{3,n})\lesssim\frac{\log_2n}{n}.$$ Also, in \cite{OK}, page 20,  it is shown that  
$$\EX(A_{4,n})\leq 2\inf\limits_{g\in\mathcal{G}_n}\bigg[||g-g_0||^2_{2,P_{\textbf{X}}}+J_n(g)\bigg]+\frac{1}{n}.$$
As the derivation of the above bound only uses the definition of the infimum of a given set, it also holds for any non-negaitve penalty $J_n$.
The remaining part $A_{2,n}$ is estimated using Theorem \ref{t1}. Namely, for $g\in\mathcal{G}_n$ and $\textbf{Z}:=(\textbf{X}, Y)$ defining $\Delta(g)(\textbf{Z}):=(Y^\dagger-g(\textbf{X}))^2-(Y^\dagger-g^\dagger(\textbf{X}))^2$, we have that for $t>0$  
\begin{equation}\label{A2}
\mathbb{P}(A_{2,n}>t)\leq\mathbb{P}\bigg(\sup\limits_{g\in\mathcal{G}_n}\frac{(\textrm{P}-\textrm{P}_n)\Delta(g)(\textbf{Z})}{t+2J_n(g)+\textrm{P}\Delta(g)(\textbf{Z})}\geq\frac{1}{2}\bigg)\leq \sum\limits_{j=0}^\infty\mathbb{P}\bigg(\sup\limits_{g\in\mathcal{G}_{n,j,t}}\frac{(\textrm{P}-\textrm{P}_n)\Delta(g)(\textbf{Z})}{2^jt+\textrm{P}\Delta(g)(\textbf{Z})}\geq\frac{1}{2}\bigg),
\end{equation}
where $$\mathcal{G}_{n,j,t}:=\bigg\{g\in\mathcal{G}_n: 2^{j-1}\mathbb{I}(j\neq 0)t\leq J_n(g)\leq 2^jt\bigg\}.$$
In order to bound the last expression of \eqref{A2} we need to check that for sufficiently large $n$ the conditions of Theorem \ref{t1} are satisfied for the classes 
$$\mathcal{\widetilde{G}}_{n,j,t}:=\bigg\{\Delta(g):[0,1]^d\times\mathbb{R}\mapsto\mathbb{R}: g\in\mathcal{G}_{n,j,t} \bigg\}.$$
First, we have (\cite{OK}, page 19) that for $n$ sufficiently large and for $\tilde{g}\in\mathcal{\widetilde{G}}_{n,j,t}$
$$\|\tilde{g}\|_\infty\leq8K_n^2 \quad \textrm{and}\quad \EX\{\tilde{g}(Z)^2\}\leq16K_n^2\EX\{\tilde{g}(Z)\}.$$
Thus,  for all sufficiently large $n$, the condition \eqref{cond1} holds for the class $\mathcal{\widetilde{G}}_{n,j,t}$ with $\omega=1/2,$ $t^\star\geq t_n$ and $t_n$ satisfying \eqref{t}.
To check the condition \eqref{cond2}, it is shown in \cite{OK}, page 19, that 
$$\mathcal{N}(u, \mathcal{\widetilde{G}}_{n,j,t}, \|\cdot\|_\infty)\leq\mathcal{N}(u/(4K_n), \mathcal{{G}}_{n,j,t}, \|\cdot\|_\infty).$$
Hence, from the condition \eqref{i} of the lemma we get that for any $t\geq t_n$ and any $\delta\geq 2^jt/8$
$$\int_{\delta/(2^{11}K_n^2)}^{\sqrt{\delta}}\log_2^{1/2}\mathcal{N}(u, \mathcal{\widetilde{G}}_{n,j,t}, \|\cdot\|_\infty)du\leq \int_{\delta/(2^{11}K_n^2)}^{\sqrt{\delta}}\log_2^{1/2}\mathcal{N}\bigg(\frac{u}{4K_n},\mathcal{G}_{n,j,t}, \|\cdot\|_\infty\bigg)du\leq\frac{c\delta\sqrt{n}}{\log_2 n}.$$
Thus, applying Theorem \ref{t1} with $t^\star=2^jt$ and with $\omega=1/2$, we get that for $n$ sufficiently large and for $t\geq t_n$,

$$\mathbb{P}(A_{2,n}>t)\lesssim\sum\limits_{j=0}^\infty\exp\bigg(-\frac{c2^jnt}{\log_2n}\bigg)\lesssim\exp\bigg(-\frac{cnt}{\log_2n}\bigg)$$
for some constant $c$ independent of $n$.
Hence,
$$\EX(A_{2,n})\leq 2t_n+\int_{2t_n}^\infty\mathbb{P}(A_{2,n}>t)dt\lesssim t_n+\frac{\log_2n}{n}\exp\bigg(-\frac{cnt_n}{\log_2n}\bigg)\lesssim t_n+\frac{\log_2n}{n}\lesssim t_n,$$
where the last inequality follows from the condition \eqref{t}.

\section*{Acknowledgement} The authors would like to thank the Associate Editor and the Reviewer for valuable and constructive comments which helped us to improve the manuscript. The research of Aleksandr Beknazaryan has been partially supported by the NWO Vidi grant: ``\textit{Statistical foundation for multilayer neural networks}''.  The research of Hailin Sang is partially supported by the Simons Foundation Grant 586789, USA.

\end{document}